%% file: Arx.tex
\documentclass[final, times]{elsarticle}
\usepackage{hyperref}
\journal{Journal of \LaTeX\ Templates}

\usepackage{amsthm}
\usepackage{graphicx}
\usepackage{amsmath}
\usepackage{mathtools}
\usepackage{enumitem}

\usepackage[ruled,linesnumbered,resetcount]{algorithm2e}
\usepackage{algpseudocode}
\usepackage{booktabs}
\usepackage[table]{xcolor}
\usepackage{pdflscape}
\usepackage{longtable}
\usepackage{multirow}
\usepackage{threeparttablex}
\usepackage{graphicx}
\usepackage[compatibility=false]{caption}
\usepackage[labelformat=simple]{subcaption}
\usepackage[numbers]{natbib}
\usepackage{url}
\usepackage{rotating}

\usepackage{hyperref}
\usepackage{lipsum}
\usepackage{import}
\usepackage[acronym]{glossaries}
\makeglossaries

\usepackage{graphicx}
\usepackage{amsmath}
\usepackage{mathtools}
\usepackage[ruled,linesnumbered,resetcount]{algorithm2e}

\usepackage{pdfpages}
\usepackage{import}
\usepackage{multicol}
\usepackage{tabularx}
\usepackage[draft]{changes}

\hypersetup{
	bookmarks=false,
	colorlinks=true,       % false: boxed links; true: colored links
	linkcolor=blue,          % color of internal links (change box color with linkbordercolor)
	citecolor=green,        % color of links to bibliography
	filecolor=magenta,      % color of file links
	urlcolor=cyan         
}

\loadglsentries{Glossaries}

\newcommand{\scalefigure}{0.20}

\def\BibTeX{{\rm B\kern-.05em{\sc i\kern-.025em b}\kern-.08em
    T\kern-.1667em\lower.7ex\hbox{E}\kern-.125emX}}

\begin{document}
%=====-------------------Table of Notes and Comments--------------------======
%\newpage
%\listofchanges[style=<list|summary>]
%\newpage

\begin{frontmatter}
	\title{A Multifactorial Optimization Paradigm for Linkage Tree Genetic Algorithm}
	%\tnotetext[mytitlenote]{Fully documented templates are available in the elsarticle package on \href{http://www.ctan.org/tex-archive/macros/latex/contrib/elsarticle}{CTAN}.}
	% Group authors per affiliation:
	%% or include affiliations in footnotes:
	
	\author[httb]{Huynh Thi Thanh Binh}
	\ead{binhht@soict.hust.edu.vn}
	
	\author[pdt,d]{Pham Dinh Thanh\corref{cor1}\href{https://orcid.org/0000-0002-2550-9546 }{\includegraphics[scale=0.5]{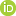}}}
	\ead{thanhpd05@gmail.com}	
	
	%% or include affiliations in footnotes:
	\author[httb]{Tran Ba Trung}%\corref{cor1}}
	\ead{batrung97@gmail.com}
	
	%% or include affiliations in footnotes:
	\author[httb]{Le Cong Thanh}%\corref{cor1}}
	\ead{thanhcls1316@gmail.com}
		
	%% or include affiliations in footnotes:
	\author[httb]{Le Minh Hai Phong}%\corref{cor1}}
	\ead{aquariuslee99@gmail.com}

	\author[c]{Ananthram Swami}%\corref{cor1}}
	\ead{ananthram.swami.civ@mail.mil}
	
	\author[d]{Bui Thu Lam}%\corref{cor1}}
	\ead{lambt@lqdtu.edu.vn}
	
	\cortext[cor1]{Corresponding author.}
	
	\address[httb]{School of Information and Communication Technology, Hanoi University of Science and Technology, Vietnam}
	\address[pdt]{Faculty of Mathematics - Physics - Informatics, Taybac University, Vietnam}
	\address[c]{United States Army Research Laboratory, USA}
	\address[d]{Le Quy Don Technical University, Vietnam}

	\begin{abstract}
		\gls{ltga} is an effective \gls{ea} to solve complex problems using the linkage information between problem variables. \gls{ltga} performs well in various kinds of  single-task optimization and yields promising results in comparison with the canonical genetic algorithm. However, \gls{ltga} is an unsuitable method for dealing with multi-task optimization problems. On the other hand, \gls{mfo} can simultaneously solve independent optimization problems, which are encoded in a unified representation to take advantage of the process of knowledge transfer. In this paper, we introduce \gls{mfltga} by combining the main features of both \gls{ltga} and \gls{mfo}. \gls{mfltga} is able to tackle multiple optimization tasks at the same time, each task learns the dependency between problem variables from the shared representation. This knowledge serves to determine the high-quality partial solutions for supporting other tasks in exploring the search space. Moreover, \gls{mfltga} speeds up convergence because of knowledge transfer of relevant problems. We demonstrate the effectiveness of the proposed algorithm on two benchmark problems: \glsentrylong{clustp} and \glsentrylong{dtf}. In comparison to \gls{ltga} and existing methods, \gls{mfltga} outperforms in quality of the solution or in computation time.
	\end{abstract}
	
	\begin{keyword}
		Linkage Tree Genetic Algorithm \sep Multifactorial Optimization \sep Linkage models, Clustered Shortest-Path Tree Problem \sep Evolutionary Algorithm.
	\end{keyword}	
\end{frontmatter}

%\linenumbers
\glsresetall

%%================------------------Sec_Introduction-------------------=============
\section{Introduction}
\label{Sec_Introduction}
\input{Sections/Sec_Introduction}

%%================------------------Sec_Related_Works-------------------=============
\section{Related works}
\label{Sec_Related_Works}
\input{Sections/Sec_Related_works}

%%================------------------Sec_Intro_MFO & LTGA-------------------=============
%\section{Multifactorial Optimizations and Linkage Tree Genetic Algorithm}
\section{Preliminaries}
\label{Sec_Intro_MFO_LTGA}

This section provides a brief background of the Multifactorial Optimization paradigm and the Linkage Tree Genetic Algorithm.
\subsection{Multifactorial Optimizations}
\input{Sections/Sec_Intro_MFO} \label{Subsec_Intro_MFO}

%%================------------------Sec_Intro_MFO-------------------=============
\subsection{Linkage Tree Genetic Algorithm}
\label{subsec_Intro_LTGA}
\input{Sections/Sec_Intro_LTGA}

%%================------------------Sec_Proposed_Algorithm-------------------=============
\section{Multifactorial Optimization with Linkage Tree Genetic Algorithm}
\label{Sec_Proposed_Algorithm}
\input{Sections/Sec_Proposed_Algorithm}

%%================------------------Sec_Computational_results-------------------=============
\section{Simulation results}
\label{Sec_Computational_results}
\input{Sections/Sec_Computational_Results}

%%%------------------------------------Sec_Conclusion---------------------------------------------%%%
\section{Conclusion}
\label{Sec_Conclusion}
\input{Sections/Sec_Conclusion}

%%%------------------------------------acknowledgements---------------------------------------------%%%
%\section*{Acknowledgment} 
%\label{Acknowledgment}
%\input{Sec_Acknowledgment}	

\bibliography{references}

%\clearpage
%\onecolumn
\begin{landscape}
%	
\input{Table_Data/ResultsType1}

\input{Table_Data/ResultsType5}

\input{Table_Data/ResultsType6}	
\end{landscape}

%\vspace{12pt}
%\color{red}

\end{document}

%% file: Glossaries.tex
\newacronym{ga}{GA}{Genetic Algorithm}
\newacronym{ea}{EA}{Evolutionary Algorithm}
\newacronym{mfo}{MFO}{Multifactorial Optimization}
\newacronym{mfea}{MFEA}{Multifactorial Evolutionary Algorithm}
\newacronym{clustp}{CluSPT}{Clustered Shortest-Path Tree Problem}
\newacronym{sto}{STO}{single-tasking of Genetic Algorithm}
\newacronym{clumrct}{CluMRCT}{Minimum Routing Cost Clustered Tree Problem}
\newacronym{clutsp}{CluTSP}{Clustered Traveling Salesman Problem}
\newacronym{ftsp}{FTSP}{Family Traveling Salesman Problem}
\newacronym{gtsp}{GTSP}{Generalized Traveling Salesman Problem}
\newacronym{gp}{GP}{Genetic Programming}
\newacronym{interclumrct}{InterCluMRCT}{Minimum Inter-cluster Routing Cost Clustered Tree Problem}
\newacronym{grasp}{GRASP}{Greedy Randomized Adaptive Search Procedure}
\newacronym{ant}{ACO}{Ant Colony Optimization}

\newacronym{tsp}{TSP}{Traveling Salesman Problem}
\newacronym{steinertp}{STP}{Steiner Tree Problem}
\newacronym{clusteinertp}{CluSteinerTP}{Clustered Steiner Tree Problem}
\newacronym{uss}{USS}{Unified Search Space}
\newacronym{de}{DE}{Differential Evolution Algorithm}
\newacronym{pso}{PSO}{Particles Warm Optimization Algorithm}
\newacronym{rmp}{rmp}{Probability of Random Mating}
\newacronym{vrp}{VRP}{Vehicle Routing Problem}
\newacronym{cvrp}{CVRP}{Capacitated Vehicle Routing Problem}
\newacronym{ltga}{LTGA}{Linkage Tree Genetic Algorithm}
\newacronym{mfed}{MFEA}{Multifactorial evolutionary algorithm}
\newacronym{rga}{RGA}{Randomized Greedy Algorithm}

%----------------------This Paper--------------------------
\newacronym{mfltga}{MF-LTGA}{Multifactorial Linkage Tree Genetic Algorithm}

\newacronym{nrga}{HBRGA}{Heuristic Based on Randomized Greedy Algorithm}
\newacronym{spta}{SPTA}{Shortest Path Tree Algorithm}

\newacronym{dtf}{DTF}{Deceptive Trap Function}
\newacronym{mk-dtf}{mk-DTF}{Deceptive mk-Trap Function}
\newacronym{scm}{SCM}{Swap-Change Mutation operator}

%% file: Sections/Sec_Introduction.tex
%\glsreset{ltga}
\glsentrylong{ltga} has been shown to scale excellently on a variety of discrete, Cartesian-space, optimization problems~\cite{bosman_expanding_2016, bouter_exploiting_2017}. \gls{ltga} determines the linkages between problem variables in the population, then clusters relevant variables to build a linkage tree. In each generation, the linkage tree is used to create crossover masks to prevent disruption between high-quality linked genes, and different partial structures of two good parent solutions can be juxtaposed to construct a new good solution. \gls{ltga} performs well in various problems: Permutation Flowshop Scheduling Problem~\cite{bouter_exploiting_2017,aalvanger_heuristics_2018}, Nearest-neighbor NK landscapes~\cite{thierens_evolvability_2012}, MAX-SAT problem~\cite{de_bokx_search_2015,sadowski_usefulness_2013}, Deceptive trap function~\cite{thierens_the_linkage_tree_genetic_algorithm_2010}, Multidimensional Knapsack Problem~\cite{martins_performance_2014}, etc. and outperforms traditional \gls{ga}. However, the linkage tree is built from a single combinatorial optimization problem without transferred knowledge from other relevant problems.

Combinatorial optimization problems in real-life like Jobs scheduling, Cloud computing etc. require solving many tasks simultaneously. Arising from the need to solve a large number of user requests in Cloud Computing, \gls{mfea} proposed by Gupta,~et~al.~\cite{gupta_multifactorial_2016} can solve multiple independent optimization problems simultaneously using a single population of solutions in the unified search space. The unified search space encompasses the shared knowledge for all of the tasks and the knowledge belonging to the particular optimization task. Transferring knowledge between different tasks occurs through adjusting and exchanging shared genetic material in the unified search space. In the process of transferring knowledge, good partial solutions of each task are used to support the others tasks. Leveraging the supportive genetic material requires calculating the commonality between all tasks for effective knowledge transfer.

Inspired by the idea of \gls{mfo} and \gls{ltga}, we exploit the advantages of crossover mechanism of \gls{ltga} and the implicit genetic transferring of \gls{mfo}.  We adopt the idea that many problems are represented in the unified search space, building linkage tree for each task exploits the knowledge from the other tasks. In particular, the linkage tree indicates the distances between problem variables. These variables corresponding to each problem are the partial structure of individuals in the shared representation. The distance between two sets of variables indicates the dependence between them, calculated. 

In this paper, we introduce \gls{mfltga} by combing the main features of \gls{ltga} and \gls{mfo}: Linkage tree is used to determine the relationship between problems variables, which is used to leverage shared information among optimization problems. The assortative mating step is modified to combine crossover mechanism of \gls{ltga} and vertical cultural transmission of \gls{mfea}: A linkage tree is selected corresponding to a particular task then the crossover operator is applied to generate new offspring based on that linkage tree. The vertical cultural transmission in \gls{mfltga} serves to transfer the phenotype of parents to their offspring.

The effectiveness of \gls{mfltga} is shown in comparison to \gls{ltga} and existing algorithms on the canonical \gls{clustp}~\cite{demidio_clustered_2016} and \gls{dtf}. The results indicate that \gls{mfltga} is superior to \gls{ltga} in computation time, and quality of the solution.

\noindent The main contributions of this paper are:
\begin{itemize}
	\item We propose a mechanism combining key features of \gls{ltga} and \gls{mfo}, which we call \gls{mfltga}.
	\item We modify process for building the linkage tree based on unified search space to exploit both information between problem variables and transfer knowledge of independent tasks.
	\item We introduce an assortative mating mechanism to enhance the compatibility between the main features of \gls{mfo} and \gls{ltga}.
	\item We propose a crossover operator to keep the key advantages of \gls{mfo} and \gls{ltga} as well as maintain population diversity.
	\item The experimental results show that the our algorithm is more efficient than existing methods.
\end{itemize}

This paper is organized as follows. Section \ref{Sec_Related_Works} introduces related works. A background of \gls{mfo} and \gls{ltga} are briefed in Section \ref{Sec_Intro_MFO_LTGA}. Section \ref{Sec_Proposed_Algorithm} describes the \gls{mfltga} algorithm. Section \ref{Sec_Computational_results} presents and discusses experimental results. The paper concludes in Section \ref{Sec_Conclusion} with discussions of the future extension of this research. %Section \ref{Sec_Conclusion} summarizes and concludes the paper.

%The rest of this paper is organized as follows. Section II introduced related works. The proposed MFEA for the CSTP is elaborated in section III. Section IV explains the setup of our experiments and reports the computed results. The paper concludes in section V with discussions on the future extension of this research.

%% file: Sections/Sec_Related_works.tex
\glsresetall

\gls{ltga} was introduced by Thierens~\cite{thierens_linkage_2010} and is one of the newest variants of \gls{ea}~\cite{agoston_eiben_2003,back_evolutionary_1996}. \gls{ltga} learns the linkage between the problem variables and groups of variables by building a hierarchical clustered tree using a proximity distance metric. In each generation, \gls{ltga} builds a linkage tree and then uses that tree to generate new offspring. \gls{ltga} has been successfully applied to various types of problems which we review next.

In~\cite{bosman_expanding_2016}, \gls{ltga} was proposed to solve permutation optimization problems by employing the random key encoding of permutations. To evaluate the dependency between two variables, two factors were proposed: the first factor, called relative-ordering information, focuses on the order of two genes while the second factor, called adjacency information, focuses on the proximity of the two genes. 

In~\cite{thierens_linkage_2010}, \gls{ltga} is applied to solve the deceptive $mk-trap$ function. The authors use the mutual information for evaluating the dependency between variables and build the linkage tree of a population of solutions.

In~\cite{goldman_linkage_2012}, Goldman, ~et~al. introduced a benchmark problem, deceptive step trap problem for testing \gls{ltga}. To reduce the time complexity when calculating the entropy between all possible clusters, the authors also proposed the linkage between clusters. Instead of finding the entropy of an entire cluster, the new measure only finds the entropy between all pairs of independent problem variables in the population. 

To improve the convergence of \gls{ltga}, Bosman,~et~al.~\cite{bosman_linkage_2012} proposed Forced Improvements which is used when a solution cannot be improved. A different linkage model, Linkage Neighbors (LN) was also proposed. An advantage of the LN model compared to the Linkage Tree
model is that it is well-suited to represent overlapping linkage relations.

Recently, the concept of \gls{mfo}~\cite{ong_evolutionary_2016, trung_multifactorial_2019} has been introduced in~\cite{gupta_multifactorial_2016} as a new optimization paradigm toward evolutionary multitasking with a single population of individuals. In contrast to traditional evolutionary search paradigm, \gls{mfo} conducts evolutionary search concurrently on multiple search spaces corresponding to the optimization problems, each possessing a unique function landscape. The efficacy of \gls{mfo} has been confirmed by a set of continuous and combinatorial optimization problems in~\cite{bosman_measures_2012,bouter_exploiting_2017,thanh_efficient_2020}.

\gls{mfo} has been applied to various algorithms which we review next. Feng,~et~al.~\cite{feng2017empirical} proposed two new mechanisms for combining \gls{mfea} with \gls{pso} (called MFPSO), \gls{de} (called MFDE). In the new algorithms, the authors designed new assortative mating schemes while the other components such as unified individual representation, vertical cultural transmission, etc., are kept the same as in the original \gls{mfea}. %In MFPSO, with probability value $rmp$, the velocity particle i is updated by following the global best solution having skill factor different from that of this particle. Otherwise, the velocity is updated as original \gls{pso}.

Xie,~et al.~\cite{xie_enhancing_2016} introduced a hybrid algorithm combining \gls{mfea} and \gls{pso} (call HMFEA) in which \gls{pso} plays the role of local search in the \gls{mfea}. A difference between HMFEA and the original \gls{mfea} is that the \gls{pso} is added after genetic operation of \gls{mfea} and applied to the intermediate-pop in each generation. To adjust dynamically the velocity and guarantee that the convergence velocity is not too fast, an adaptive variation adjustment factor $g_\alpha$ is proposed. The factor $g_\alpha$ is used to control the velocity of each particle.

In~\cite{wen_learning_2016}, Wen and Ting combine the \gls{mfea} with \gls{gp} for learning an ensemble of decision trees. In this algorithm, each task is associated with one run of \gls{gp}. To generate diverse decision trees, their algorithm further scrambles the dataset for each task by randomly mapping the feature indexes. The tasks will then work on the dataset with different feature sequences.

%\glsreset{gp}
Zhong.~et~al.~\cite{zhong_multifactorial_2018} proposed a multifactorial \gls{gp} (MFGP) paradigm toward evolutionary multitasking \gls{gp}. MFGP consists of a novel scalable chromosome encoding scheme and new evolutionary mechanisms for \gls{mfo} based on self-learning gene expression programming.

Although \gls{mfo} and \gls{ltga} were developed for solving various type of problems, there have been no studies that combine the strengths of \gls{mfo} and \gls{ltga} into a new algorithm. Therefore, this paper proposes mechanisms to take the advantages of both \gls{mfo} and \gls{ltga} into a new algorithm. The  experimental results demonstrate the effectiveness of the new algorithm.

%% file: Sections/Sec_Intro_MFO.tex
In~\cite{gupta_multifactorial_2016}, Gupta et al. introduced \acrlong{mfo} as an evolutionary multi-tasking paradigm that optimizes multiple tasks simultaneously. Unlike traditional methods, \gls{mfo} solves multiple tasks within only a single task. To achieve this, individuals are represented in unified search space and \gls{mfo} calculates the skill of individual and splits the population into different groups: each individual is placed in the group corresponding to the task it performs best. The ability to solve problems in multitasking environments not only allows \gls{mfo} to utilize genetic materials created in a specific group but also useful for another task. 

To evaluate an individual, Gupta~et~al.~\cite{gupta_multifactorial_2016} define the following properties for every individual $p_i$ in population P:

\begin{itemize}
	\item \textbf{Factorial Cost}: The factorial cost $\psi^i_j$ of an individual  $p_i$ on task $T_j$ is computed by its fitness or objective value on a particular task $T_j$.
	\item \textbf{Factorial rank}: Factorial rank $r^i_{j}$ is the rank of $p_i$ on task $T_j$, relative to all other individuals in P.
	\item \textbf{Scalar Fitness}: Scalar fitness $\varphi_i$ of $p_i$ is based on its best rank over all tasks; i.e. $\varphi_i$ = 1/min\{$r^i_{1}$, $r^i_{2}$, \ldots, $r^i_{K}$\}.
	\item \textbf{Skill Factor}: Skill factor $\tau_i$ of $p_i$ is the one task, amongst all other tasks in \gls{mfo}, with which the individual is associated. This may be defined as   $\tau_i = argmin_j\{r^i_{j}\}$.
\end{itemize}

In order to calculate fitness of an individual, individuals are decoded in different tasks to obtain ``Factorial Cost”. Individuals are evaluated by its correlation with other individuals based on ``Factorial Cost” to find the most suitable task called ``Skill Factor”. %Clearly, by means of the above properties, \gls{mfo} has provided evaluation methods for individuals in multitasking environment. 

%% file: Sections/Sec_Intro_LTGA.tex
Recently, a powerful linkage-learning \gls{ea}, \gls{ltga}, was proposed by Dirk Thierens~\cite{thierens_linkage_2010}. \gls{ltga} maximizes the effectiveness of the crossover operator through discovering and exploiting the relationship between problem variables during the evolutionary searching. To store linkage information, \gls{ltga} uses an additional hierarchical tree, called linkage tree. A cluster of problem variables that \gls{ltga} believes to be linked is represented by a node in the linkage tree. In each generation, the linkage tree is rebuilt by selecting a set of solutions from the current population before determining the relationship between problem variables in that set. 

\subsubsection{Constructing Linkage Tree}
\gls{ltga} aims to identify the variables that make a dependent set, then uses an agglomerative hierarchical clustering algorithm to proceed bottom-up. Hierarchical clustering algorithm constructs Linkage information between variables, and stores it as a Linkage Tree. Each node in the Linkage Tree is a cluster of genes that are close to each other. At its first stage, the algorithm considers each gene to be a dependent cluster, before repeatedly joining the two closest clusters to create a bigger one until all genes are in the same cluster. The size of population may impact the accuracy of the information the linkage tree represents. The larger the population size, the higher the possibility of good solutions appearing in it. Therefore, the linkage tree constructed from larger population may better reflect the relations between the genes. However, for larger population, the construction of linkage tree would be more consuming in terms of computational resources and the number of evolutionary operations on each generation would be higher. Hence, it is necessary to choose an appropriate population size in order to keep a balance between the linkage information accuracy and computational resources consumption.

The details are shown in Algorithm~\ref{alg:HierarchicalClustering}:

\begin{algorithm}
	\KwIn{A set of solutions from the current population}
	\KwOut{A Linkage tree}
	\BlankLine
	Compute the proximity matrix using metric $D$.\\
	Assign each variable to a single cluster.\\
	Repeat until one cluster left:\\
	\quad Join two closest clusters $C_i$ and $C_j$ into $C_{ij}$.\\
	\quad Remove $C_i$ and $C_j$ from the proximity matrix.\\
	\quad Compute the distance between $C_{ij}$ and all clusters.\\
	\quad Add cluster $C_{ij}$ to the proximity matrix.\\
	\caption{Hierarchical Clustering}
	\label{alg:HierarchicalClustering}
\end{algorithm}

\renewcommand{\scalefigure}{0.35}
\begin{figure}[htbp]
	\centering
	\includegraphics[scale=\scalefigure]{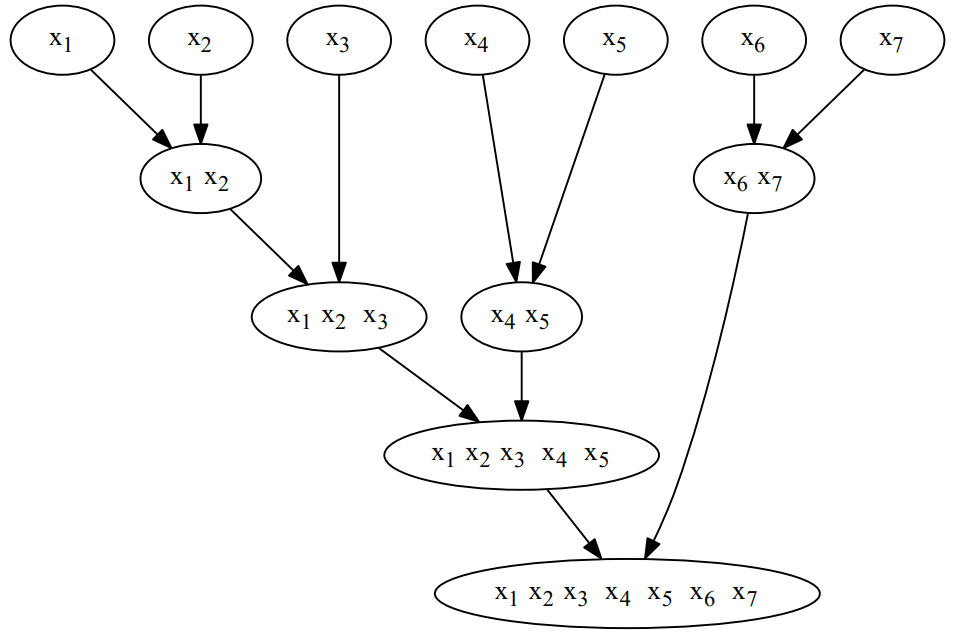}
	\caption{An example of hierarchical clustering algorithm on 7 genes.}
	\label{fig:hierarchicalclustering}
\end{figure}

An example of hierarchical clustering is shown in Figure~\ref{fig:hierarchicalclustering}: The first two closest genes $x_1$ and $x_2$ are joined into a cluster $x_1 x_2$, clusters $x_1$, $x_2$  are removed from the proximity matrix. In the next iteration, \gls{ltga} considers the distances between the new cluster $x_1 x_2$ and the other clusters, then combines the closest pair of clusters from the current population. After each generation, \gls{ltga} rebuilds the linkage tree from current population. %\highlight[comment={did not understand this;i think it can be deleted}]{the larger the number of individual, the more accurately the linkage tree represents}.

\subsubsection{Crossover operator}
 Each cluster in the Linkage Tree is used as a crossover mask, the variables in a cluster are swapped between parent pair to generate two new offspring. If one of the offspring is better than its parents, then those offspring become parents for the next crossover for the remaining crossover masks. \gls{ltga} performs operations through $2l-1$ clusters in Linkage Tree, the order of visiting clusters to perform crossover operations impacts the quality of the final solution. 

\begin{figure}
	\centering
	\includegraphics[scale=0.45]{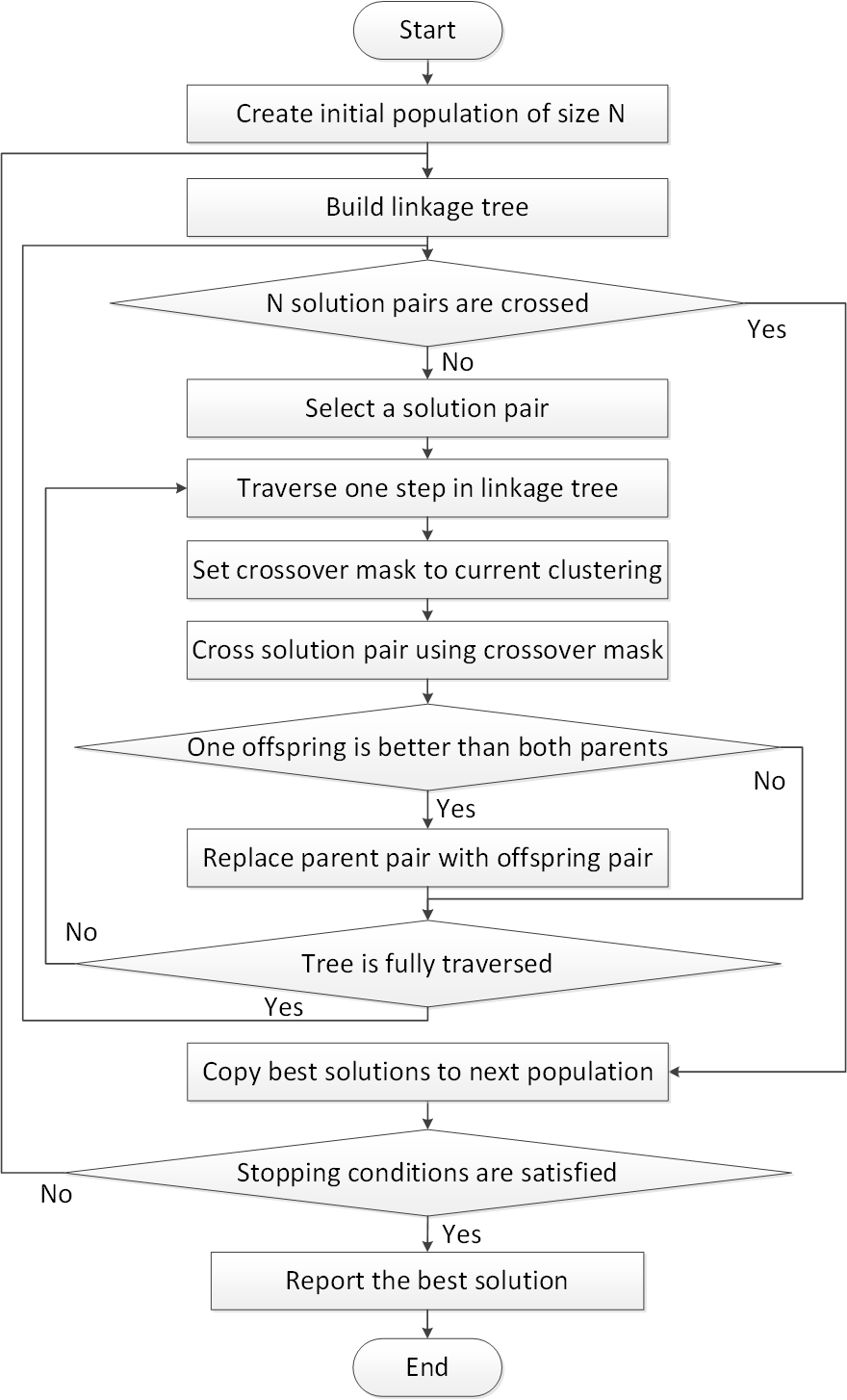}
	\caption{The outline of Linkage Tree Genetic Algorithm}
	\label{fig:WorkFlow_of_LTGA}
\end{figure}

The outline of Linkage Tree Genetic Algorithm is presented in Figure~\ref{fig:WorkFlow_of_LTGA}.

%% file: Sections/Sec_Proposed_Algorithm.tex
\glsresetall
%This section introduces a new encoding scheme for individuals as well as genetic operators of MFEA for CSTP problems.

In this section, we introduce the combination of  \gls{mfo} and \gls{ltga} which we call \gls{mfltga}. 

\gls{mfo} is designed for conducting multitasking with the exchange of information among individuals through two key components: assortative mating and vertical cultural transmission. In a standard genetic algorithm and \gls{mfo}, the solution representation and the crossover operator need to be designed to achieve good solution. However, this design will be difficult to achieve if there is insufficient domain knowledge. Different from the genetic algorithm, \gls{ltga} possesses unique solution reproduction and update operations through linkage models which learn the relationship between the problems variable through estimation of distribution. 

\begin{center}
	\begin{figure}
		\includegraphics[scale=0.45]{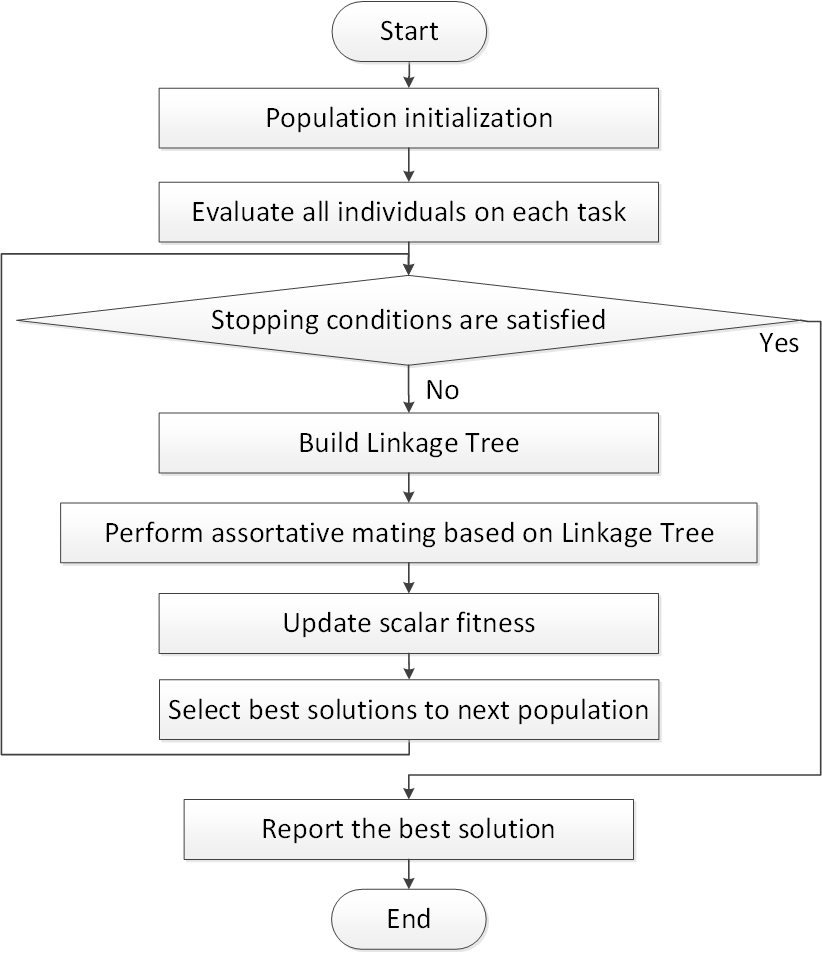}
		\caption{The scheme of Multifactorial Linkage Tree Genetic Algorithm}
		\label{fig:WorkFlow_of_MF_LTGA}
	\end{figure}
\end{center}

\gls{mfltga} is our proposal by combining \gls{ltga} and \gls{mfo} in order to capture the advantages of both algorithms to improve the quality of the solution. To hybridize \gls{mfo} and \gls{ltga}, new assortative mating schemes are required. In addition, some operators like unified individual representation, vertical cultural transmission, etc., also need to change to adapt to \gls{ltga}. The workflow in Figure~\ref{fig:WorkFlow_of_MF_LTGA} describes the outline of our proposed algorithm, in which we maily focus on the two steps: \textit{build linkage tree} and \textit{perform associative mating based on linkage tree}. \gls{mfltga} start with a initial population of individuals which is presented in a unified search space. \textit{The assortative mating} of \gls{mfltga} serves as the genetic operator to reproduce next generation as well as \gls{mfea}. However, \textit{the assortative mating} of \gls{mfltga} is performed based on \textit{linkage tree} which learns a probabilistic model of the current population of solutions. In addition, unlike \gls{mfea}, \textit{vertical cultural transmission} is determined in assortative mating because it depends on tree selection. The pseudo code of \gls{mfltga} is presented in Algorithm~\ref{alg:PseudoCodeMF-LTGA}. In what follows, the design of the \gls{mfltga} is detailed. 

\begin{algorithm}
	\textbf{Generate} an initial population of size $n$\;
	\textbf{Evaluate} all individuals on every task in the multi-tasking environment, and obtain the skill factor of each individual\;
	$t \leftarrow 0$\;
	\While{stopping conditions are not satisfied}
	{	
		\textbf{Build} linkage tree \Comment{Refer to Algorithm~\ref{alg:BuildLinkageModel}}\;
		\textbf{Perform} \textit{assortative mating} based on linkage model on \textit{current-pop} to generate an \textit{intermediate-pop} \Comment{Refer to Algorithm~\ref{alg:AssortativeMating}}\;
		\textbf{Update} the scalar fitness of all the individuals in \textit{intermediate-pop} \Comment{Refer subsection~\ref{Subsec_Intro_MFO}}\;
		\textbf{Select} the fittest individuals from \textit{intermediate-pop} to form the next \textit{generation}\;
		$t \leftarrow t + 1$\;
	}
	\glsreset{mfltga}
	\caption{\gls{mfltga}}
	\label{alg:PseudoCodeMF-LTGA}
	\BlankLine
\end{algorithm}

%%=====-----------------------Build Linkage Tree Model-----------------------====
\subsection{Linkage Tree Building}
A key strength of \gls{ltga} is its ability to learn the relationship between the problem variables. To maintain this strength when applied to a multi-tasking environment, either a linkage tree is built for all tasks, or linkage trees are built separately for each task. Building only a single tree for all tasks can not provide the dependency between the variables because the relationship between two variables in one task might be different from that in another task. Therefore, this paper applies the second approach. The pseudo code of the building linkage tree in \gls{mfltga} is given in Algorithm~\ref{alg:BuildLinkageModel}. 

In Algorithm~\ref{alg:BuildLinkageModel}, for each task, we firstly generate a selected population including individuals whose Skill Factor is in that task. Next, each individual in the selected population is decoded to a solution for this task which will be added to a population, called task-population. Finally, linkage tree is built based on task-population in the same way in~\cite{thierens_linkage_2010}. 
%Tree selection on crossover operator will be referred in part .

\begin{algorithm}
	\KwIn{$P$: A population of individuals in unified search space; $k$: Number of tasks.}
	\KwOut{Linkage tree $T_i$ for task $i, i=1,\ldots,k$.}
	\BlankLine
	\ForEach{task $i$}
	{	
		\textbf{Choose} individuals from $P$ which have skill factor of task $i$ to generate \textit{selected-population} ($C$)\; %>>>>>>>>>>chỗ này chưa ổn
		\textbf{Decode} all individuals of \textit{selected-population} ($C$) to generate \textit{task-population} ($P_{i}$)\;
		\textbf{Build} the linkage tree (${T_{i}}$)  based on \textit{task-population} ($P_{i}$);
	}
	\Return $T_i, i=1,\ldots,k$
	\caption{Linkage tree building in \gls{mfltga}}
	\label{alg:BuildLinkageModel}
\end{algorithm}

\subsection{Assortative Mating}
The pseudo code of the assortative mating in \gls{mfltga} is given in Algorithm~\ref{alg:AssortativeMating}. Firstly, current-population is partitioned into pairs of individuals which are considered as parents. Next, with each pairs of parents, we select randomly a single task for evaluation, because evaluating every individual for every problem being solved will often be computationally too expensive. However, comparisons and evaluations only on selected task may lead to a loss on good individuals of unselected tasks. These individual that has unselected skill factor might be an outstanding solution for that particular task and could produce good offspring on that task. Therefore, we need to create a backup population which contains the individuals that does not have their skill factor tasks selected. Next, the pair of parents will have crossover operator applied based on the tree of the selected task to generate offspring. As a result, the offspring imitate selected task, so that \textit{vertical cultural transmission} is integrated into the assortative mating. Finally, offspring-pop and backup-pop are concatenated to form an intermediate-pop. 

\begin{algorithm}
	\KwIn{\begin{itemize}
			\item $P$: A population of individuals in unified search space;
			\item $k$: Number of tasks;
			\item Linkage tree $T_i$ for task $i, i=1,\ldots,k$.
		\end{itemize}
		  }
	\KwOut{A new population of individuals in unified search space} %$P'$}
	\BlankLine
	Build a \textit{set of selected parents} ($S$) by randomly partitioning population ($P$) into pairs of individuals\;
	\ForEach{pair ($p_i, p_j$) in $S$}
	{	
		$\tau_i \leftarrow$ skill factor of $p_i$\;
		$\tau_j \leftarrow$ skill factor of $p_j$\;
		$\tau \gets \tau_i$\;
		\If{$\tau_i \neq \tau_j$}
		{
			$\tau \gets $ A random skill factor from $\{\tau_i,\tau_j\}$\;
			\If{$\tau = \tau_i$}
			{
				Add $p_j$ to \textit{backup population} ($B$)\;
			}
			\Else
			{
				Add $p_i$ to \textit{backup population} ($B$)\;
			}
			
		}
		Parents $p_i$ and $p_j$ crossover in task $\tau$ based on $T_{\tau}$ to generate two offspring $o_i$ and $o_j$ \Comment{Refer to Algorithm~\ref{alg:CrossoverOperator}}\;
		Two offspring $o_i$ and $o_j$ imitates selected skill factor $\tau$ \;
		Add best individual of $\{o_i, o_j\}$ to \textit{offspring population} ($O$)\;
	}
	Concatenate \textit{offspring-pop} and \textit{backup-pop} to form an \textit{intermediate-pop} ($O \cup B $)\;
	\Return \textit{intermediate-pop}\;
	\caption{Assortative Mating in \gls{mfltga}}
	\label{alg:AssortativeMating}
\end{algorithm}

%%=====-----------------------Crossover Operator-----------------------====
\subsection{Crossover Operator}
In this part, we will clarify the crossover operator based on linkage tree. The pseudo code of the crossover operator is presented in Algorithm~\ref{alg:CrossoverOperator}. In the new crossover operator, we traverse the linkage tree top-down to set the crossover mask. With each mask, parent pair is crossed using crossover mask to generate a pair of offspring which is evaluated on the selected task to compete with the parent pair. If one of the children is better than both parents then the offspring pair replaces the parent pair, and \gls{mfltga} continues to traverse the linkage tree with the new pair. If none of the two children is better than their parents, \gls{mfltga} continues its tree traversal with the parent pair. However, after some traversal in the tree, some individuals cannot be improved. Therefore, we propose a mechanism for replacing these individuals by new individuals. Whenever an individual cannot be improved, we will punish this individual; once the individual's punishment record reaches the threshold, it is replaced by a new individual. 

\begin{algorithm}
	\KwIn{\begin{itemize}
			\item Parents $p_i$ and $p_j$;
			\item $max_p$: maximum punishment;
			\item Linkage tree $T_\tau$ for task $\tau$;
		\end{itemize}
	}
	\KwOut{Two offspring $o_i$ and $o_j$;}
	\BlankLine
	change $\gets$ False\;
	$n_p \gets$ 0\;
	\ForEach{node $t_{i}$ in $T_{\tau}$}
	{
		Set crossover mask to node $t_{i}$\;
		Cross parents $p_i$ and $p_j$ using crossover mask to generate offspring $o_i$ and $o_j$\;
		
		\If{one offspring better than both parents}
		{
			Replace parents $p_i$ and $p_j$ by offspring $o_i$ and $o_j$\;
			change $\gets$ True\;
		}
	
	}
	\If{!change}
	{
	$n_p\gets n_p$ + 1\;
	\If{$n_p > max_p$}
	{
		Replace parents $p_i$ and $p_j$ by two new random individuals\;
		%				p $\gets$ new Individual
		$n_p \gets$ 0\;
	}
	change $\gets$ False\;
	}
	
	\Return Two offspring $o_i$ and $o_j$\;
	\caption{Crossover Operator in \gls{mfltga}}
	\label{alg:CrossoverOperator}
\end{algorithm}

%% file: Sections/Sec_Computational_Results.tex
%\glsentrytext, \glsentryshort or \glsentrylong
We evaluate the performance of \gls{mfltga} on two canonical problems: \gls{clustp}~\cite{demidio_clustered_2016, demidio_hardness_2019} and \gls{dtf}~\cite{thierens_evolvability_2012, thierens_linkage_2010}.  These two problems are described in detail later in this section. 

%===========----------------------Experimental setup----------------------===========
\subsection{Evaluation criteria}
We focus on the following criteria to assess the quality of the output of the algorithms.
\begin{center}
	\begin{tabular}{p{3cm}p{8cm}}
		\toprule 
		\multicolumn{2}{c}{\textbf{Criteria}} \\ 
		\midrule 
		Average (Avg)	 & Average function value over all runs. \\ 
		\addlinespace 
		Best-found (BF) & Best function value achieved over all runs. \\ 
		\addlinespace 
		Num.Opt & The number of optimal solutions are found.\\
		\addlinespace
		Num.Eval & Number of evaluations to success.\\
		\bottomrule
	\end{tabular} 
\end{center}

We compare the performance of algorithms via a normalized difference. More specifically, let $C_A$ and $C_B$ denote the performance of algorithms A and B under metric C; then the relative performance of algorithm A relative to that of algorithm B is defined as 
\begin{center}
	$PI(A,B)=\dfrac{C_B-C_A}{C_B}*100\%$
\end{center}
As examples: C could denote the cost of the best solution found by an algorithm, or the average number of evaluations needed to obtain a solution. 

%To compare the effectiveness of the two algorithms, we evaluated the difference between the average number of evaluations needed to find successfully the optimal solutions or the costs of the results obtained by the two algorithms A and B by the following formula:
%
%\begin{center}
%	$PI(A,B)=\dfrac{C_B-C_A}{C_B}*100\%$
%\end{center}
%where $C_A$ and $C_B$ denote the costs of the best solutions generated by A and B algorithms respectively or the average number of evaluations needed to find successfully the optimal solutions by algorithms A and B respectively.

To evaluate the performance of the \gls{mfltga} in solving the \gls{clustp} and the \gls{dtf}, we implemented three sets of experiments.

\begin{itemize}
	\item[$\bullet$] In the first set, the quality of the solutions obtained by the \text{C-MFEA}~\cite{ThanhPD_DungDA} and E-MFEA~\cite{ThanhPD_TrungTB} on each instance were compared with those obtained by \gls{mfltga}.	
	\item[$\bullet$] In the second set, various experiments were performed to analyze possible influencing factors. 
	\item[$\bullet$] In the third set, analyze the effective of \gls{mfltga} on instances of the Deceptive Trap Function.
\end{itemize}

This paper uses the decoding method and evolutionary operators in \cite{ThanhPD_DungDA}.

Each problem instance was evaluated 30 times for the \gls{clustp} and 10 times for the \gls{dtf} on Intel Core i7-3.60GHz, 16GB RAM computer using Windows 8 64-bit. The source codes of \gls{ltga} and \gls{mfltga} were written in the Python. 

The simulation parameters include population size = 100, number of evaluations = $10^6$, probability of random mating~=~0.5, mutation rate = 0.05 and number of tasks = 2.

%%================------------------Deceptive trap-------------------=============
\subsection{\glsentrylong{dtf}}
The Deceptive Trap Function is a well-known canonical benchmark for \gls{ltga}. With $m$-Trap Functions, the number of local optima of the deceptive trap function is $2^m-1$. A notable point for this problem is that a hillclimbing algorithm quickly becomes trapped in one of the local optima while \gls{ga} will quickly converge to the deceptive local optima~\cite{thierens_linkage_2010}.

\subsubsection{Problem formulation}
The \gls{mk-dtf} is a binary, additively decomposable function composed of $m$ trap functions $DT_i$, each defined on a separate group of $k$ bits (the total length is $l = mk$). The cost of \gls{mk-dtf} is defined as:
\begin{equation}
	f_{mk-DTF}(x_{1} \ldots x_{l})=\sum_{i=0}^{m-1}DT_i(x_{ik} \ldots x_{ik+k-1})
\end{equation}
with $x_i \in \{0,1\}$

Call $u$ the number of bits in such a group that are equal to 1:
\begin{equation}
	DT_i(x_{ik} \ldots x_{ik+k-1})=
	\begin{cases}
		k, & \text{if } u = k\\
		k-1-u, & \text{otherwise}
	\end{cases}
\end{equation}

Clearly, the array of all 1-bits is the global optimal solution of the \gls{mk-dtf} and all schemata of order less than $k$ are deceptive.

%===========----------------------Problem instances----------------------===========
%\subsubsection{Problem instances} 

%===========----------------------Experimental setup----------------------===========
\subsubsection{Experimental setup}
The \gls{mfltga} is tested on deceptive functions with trap length $k$ = 4, 5, 6. The number of blocks $m$ varies from 5 to 30 with increments of 5, the problem length thus varies from 20 to 180. The details of problem instances and parameters are presented in Table~\ref{tab:mk-dtf_instances}.

\input{Table_Data/DeceptiveTrapSetup-1}

Each problem instance was evaluated 10 times on Intel Core i7-3.60GHz, 16GB RAM computer using Windows 8 64-bit. 

%===========----------------------Experimental results----------------------===========
\subsubsection{Experimental results}
Table~\ref{tab:ResultsDeceptiveTrap} presents the results obtained by two algorithms \gls{ltga} and \gls{mfltga}. The results on this table indicate that \gls{ltga} slightly outperforms \gls{mfltga}: \gls{ltga} gets optimal solutions in all 100 tests while \gls{mfltga} only finds optimal results in 99 out of 100 tests. 

\input{Table_Data/ResultsDeceptiveTrap-2}

However, the number of evaluations needed to find successfully the optimal solutions of \gls{mfltga} is smaller than that of \gls{ltga} in all instances, and nearly 18\% fewer on average. This difference climbs up to 47.3\% on the test case $k=5, m=15$. Detailed comparisons are given in Figure~\ref{fig:DeceptiveTrap_Compare_NumEvals} in which Num.Evals denotes number of evaluations needed to find successfully the optimal solutions.

%On average number of evaluations needed to find successfully the optimal solutions, \gls{mfltga} needs about 19\% less than \gls{ltga},

\renewcommand{\scalefigure}{0.48}
\begin{figure*}[htbp]
	\centering
	\includegraphics[scale=\scalefigure]{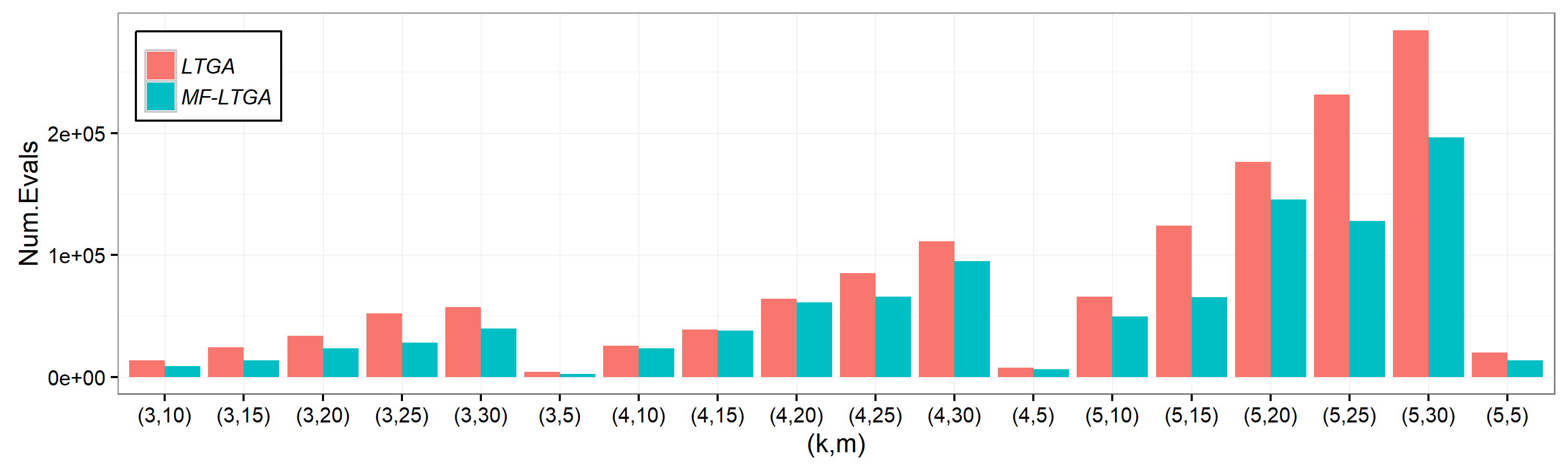}
	\caption{Comparing number of evaluations needed to find success the optimal solutions}
	\label{fig:DeceptiveTrap_Compare_NumEvals}
\end{figure*}

%===========----------------------CluSPT----------------------===========
\subsection{\glsentrylong{clustp}}
\subsubsection{Problem formulation}
In this paper, we let $G = (V, E, w)$ represent a simple, connected and undirected graph, with vertex set $V$, edge set $E$, and non-negative edge weights $w$, respectively. An edge between vertices $u$ and $v$ is denoted by $(u, v)$, and its weight is denoted by $w(u, v)$.

For a vertex subset $U$, the sub-graph of $G$ induced by $U$ is denoted by $G[U]$. A collection $C~=~\{C_i | 1 \leq i \leq k\}$ of subsets of $V$ is a partition of $V$ if the subsets are mutually disjoint and their union is exactly $V$. A path in $G$ is simple if no vertex appears more than once on the path. This paper only considers simple paths.

For a given spanning tree $T$ of $G = (V, E, w)$ and $u, v \in V$, let $d_T(u, v)$ denote the shortest path length between $u$ and $v$ on $T$.

The \gls{clustp} problem~\cite{demidio_clustered_2016, } is defined as following 
%\begin{table*}[!htb]
\begin{center}
%	\begin{tabular}{l p{6.5cm}} %<----2 cot
	\begin{tabular}{l p{9.5cm}}
		%		\hline 
		%		\multicolumn{2}{c}{\textbf{Clustered Shortest-Path Tree Problem}} \\ 
		%		\hline 
		\toprule 
		\textbf{Input}:		
		&  - A weighted undirected graph $G = (V, E, w)$.\\
		&  - Vertex set $V$ is partitioned into $k$ clusters ${C_1, C_2, . . ., C_k}$.\\
		&  - A source vertex $s \in V$.\\
		\hline
		\textbf{Output}:   	
		&  - A spanning tree $T$ of $G$.\\
		&  - Sub-graph $T[C_i] (i = 1,\ldots, k)$ is a connected graph.\\
		\hline 
		\textbf{Objective}: & $\displaystyle \sum_{v \in V} d_{T}(s,v) \rightarrow $ min\\
		%		& where $d_{T}(u, v)$ is the cost of shortest path from vertex $u$ to vertex $v$ on $T$.\\ 
		\bottomrule 
	\end{tabular}
\end{center}
%\end{table*}

\renewcommand{\scalefigure}{0.22}
\begin{figure*}[htbp]
	\centering
	\begin{subfigure}{.32\linewidth}
		\centering
		\includegraphics[scale=\scalefigure]{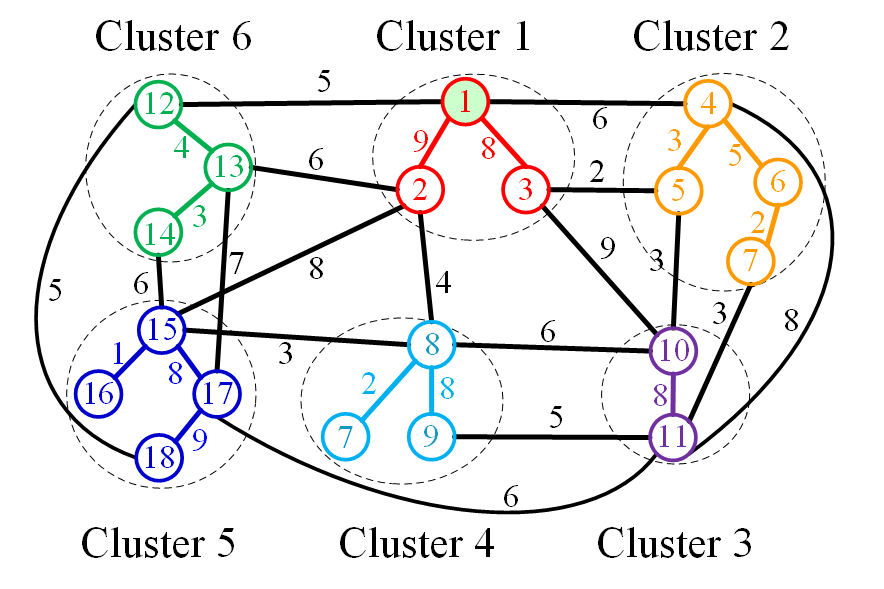}
		\caption{}
		\label{fig:InvalidSolution_a}
	\end{subfigure}
	\begin{subfigure}{.32\linewidth}
		\centering
		\includegraphics[scale=\scalefigure]{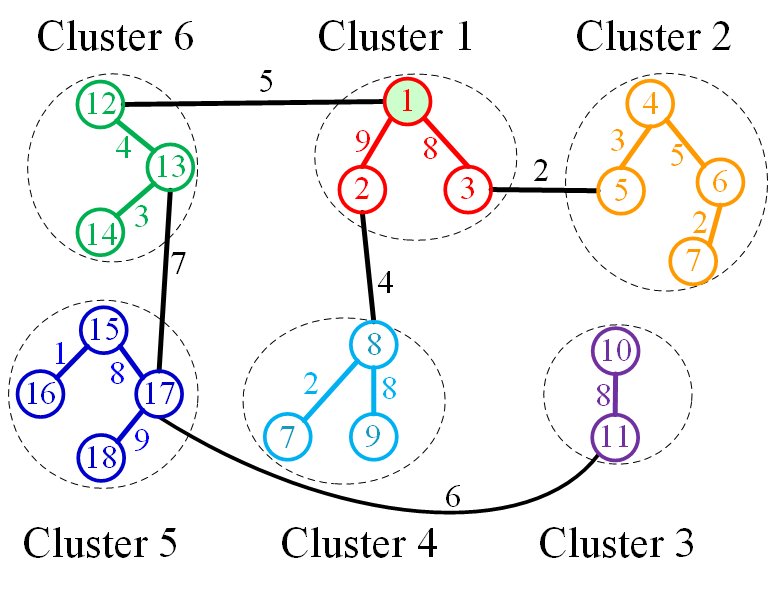}
		\caption{}
		\label{fig:InvalidSolution_b}
	\end{subfigure}
	\begin{subfigure}{.32\linewidth}
		\centering
		\includegraphics[scale=\scalefigure]{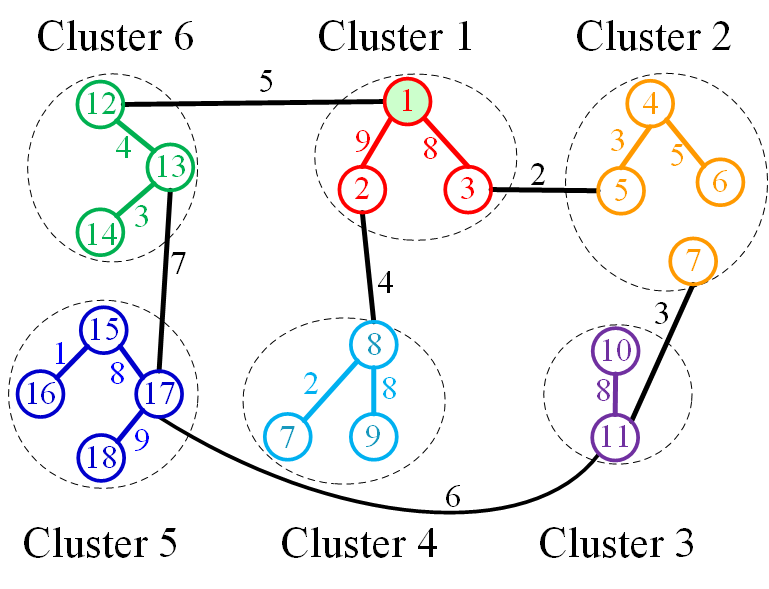}
		\caption{}
		\label{fig:InvalidSolution_c}
	\end{subfigure}
	\caption{An example of valid and  invalid solutions for the \gls{clustp}}
	\label{fig:An_example_of_invalid_solutions_in_CSTP}
\end{figure*}

Figure~\ref{fig:An_example_of_invalid_solutions_in_CSTP} illustrates the cases of valid and invalid solutions of \gls{clustp}. Figure~\ref{fig:InvalidSolution_a} shows the input graph G with 6 clusters, 18 vertices and vertex 1 as source vertex. Figure~\ref{fig:InvalidSolution_b} presents a valid solution of \gls{clustp}. In Figure~\ref{fig:InvalidSolution_c}, the vertex 6 and vertex 7 in cluster 2 are not connected, so this solution violates the second condition of the output of the \gls{clustp} problem.

%===========----------------------Problem instances----------------------===========
\subsubsection{Problem instances} 
For assessment of the proposed algorithms' performance, we created instances for \gls{clustp} from \gls{clutsp} instances \cite{helsgaun_solving_2011, mestria_grasp_2013} by adding the information of the source vertex. The main reason for building \gls{clustp} instances from \gls{clutsp} instances was that \gls{clutsp} instances have been proved to be suitable for clustered problems in general \cite{mestria_grasp_2013}.

%For evaluation of the proposed algorithms, small instances of \gls{clustp} were selected.

All tested instances are available via~\cite{Pham_Dinh_Thanh_2018_Instances}

%===========----------------------Experimental results----------------------===========
\subsubsection{Experimental results}
\paragraph{Comparison between the performance of existing algorithms and that of \gls{mfltga}}
In this section, we compare the results obtained by C-MFEA~\cite{ThanhPD_DungDA} and E-MFEA~\cite{ThanhPD_TrungTB} with those achieved by \gls{mfltga}. Tables~\ref{tab:ResultsType1},~\ref{tab:ResultsType5}~and~\ref{tab:ResultsType6} illustrate the results obtained by these algorithm on instances of Type 1, Type 5 and Type 6. In table~\ref{tab:ResultsType6}, the symbol ``-" indicates that the corresponding instances were not executed by C-MFEA. 

The results in Table~\ref{tab:ResultsType1},~\ref{tab:ResultsType5}~and~\ref{tab:ResultsType6} point out that both single-tasking (ST) and multitasking (MT) outperform both E-MFEA and C-MFEA in most test cases. In particular, both MT and ST outperformed the two existing algorithms on all test cases in Type 5. Table~\ref{tab:SummaryResults} summarizes the comparison results among E-MFEA, C-MFEA, ST and MT on the benchmarks.

\input{Table_Data/SumeryResults}

The experimental results point out that MT is also better than ST on approximately 68\% of the test cases i.e., 17 out of 26 Type 1 instances, 10 out of 14 Type 5 instances and 25 out of 36 Type 6 instances. Maximum PI(MT, ST) are 2.5\% (for Type 1), 2.7\% (for Type 5) and 3.3\% (for Type 6). 
%The PI of results obtained by C-MFEA, E-MFEA and MF-LTGA  are given in Figure xxx.

The experimental results obtained by C-MFEA, E-MFEA and \gls{mfltga} on Type 1 instances are shown in Table~\ref{tab:ResultsType1}. On this set of instances, MT outperforms ST on 13 out of 18 instances (values in red). C-MFEA and E-MFEA outperforms ST on 3 and 9 out of 18 instances (values in italics). C-MFEA and E-MFEA outperforms MT on 5 and 9 out of 18 instances respectively (values in bold).

The results on Type 6 instances are displayed in Table~\ref{tab:ResultsType6}. ST outperformed C-MFEA on 10 out of 12 instances and outperformed E-MFEA on 15 out of 20 instances. 

%====---------------------Subsubsection: Convergency----------------------==
\paragraph{Convergence trends}
We use the functions in~\cite{gupta_multifactorial_2016} for computing the normalized objectives and averaged normalized objectives, and analyze the convergence trends of the proposed algorithm. 

\renewcommand{\scalefigure}{0.9}
\begin{figure}[htbp]
	\centering
	\includegraphics[scale=\scalefigure]{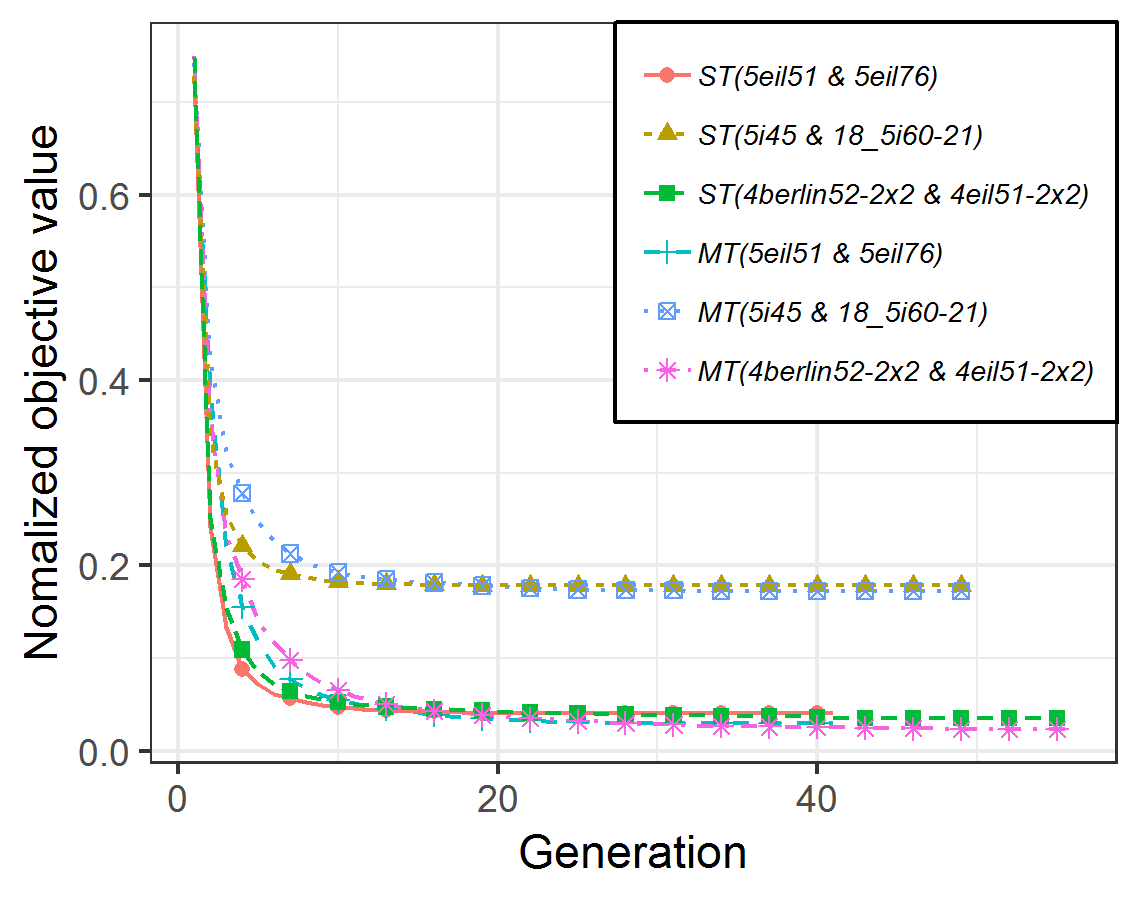}
	\caption{Convergence trends of $\tilde{f}$ in multi-tasks and serial single task for instances 10eil51 and 10eil76 in Type 1;  instances 10i60-21 and 10i65-21 in Type 5; instances 4berlin52-2x2 and 4eil51-2x2 in Type 6.}
	\label{fig:ConvergenceTrends}
\end{figure}

Figure~\ref{fig:ConvergenceTrends} illustrates the convergence trends of the ST and MT for instances 10eil51 and 10eil76 in Type 1;  instances 10i60-21 and 10i65-21 in Type 5; instances 4berlin52-2x2 and 4eil51-2x2 in Type 6. These figures point out that the convergence rate of MT is faster than that of ST in most test cases.

A notable point in Figure~\ref{fig:ConvergenceTrends} is that the numbers of evaluations of each generation are proportional to the dimensionalities of the instances. Moreover, in this experiment, the number of evaluations of each generation is a constant parameter. Due to this reason, the number of generations among instances might vary. 

\renewcommand{\scalefigure}{0.9}
\begin{figure}[htbp]
	\centering
	\begin{subfigure}{1\linewidth}
		\centering
		\includegraphics[scale=\scalefigure]{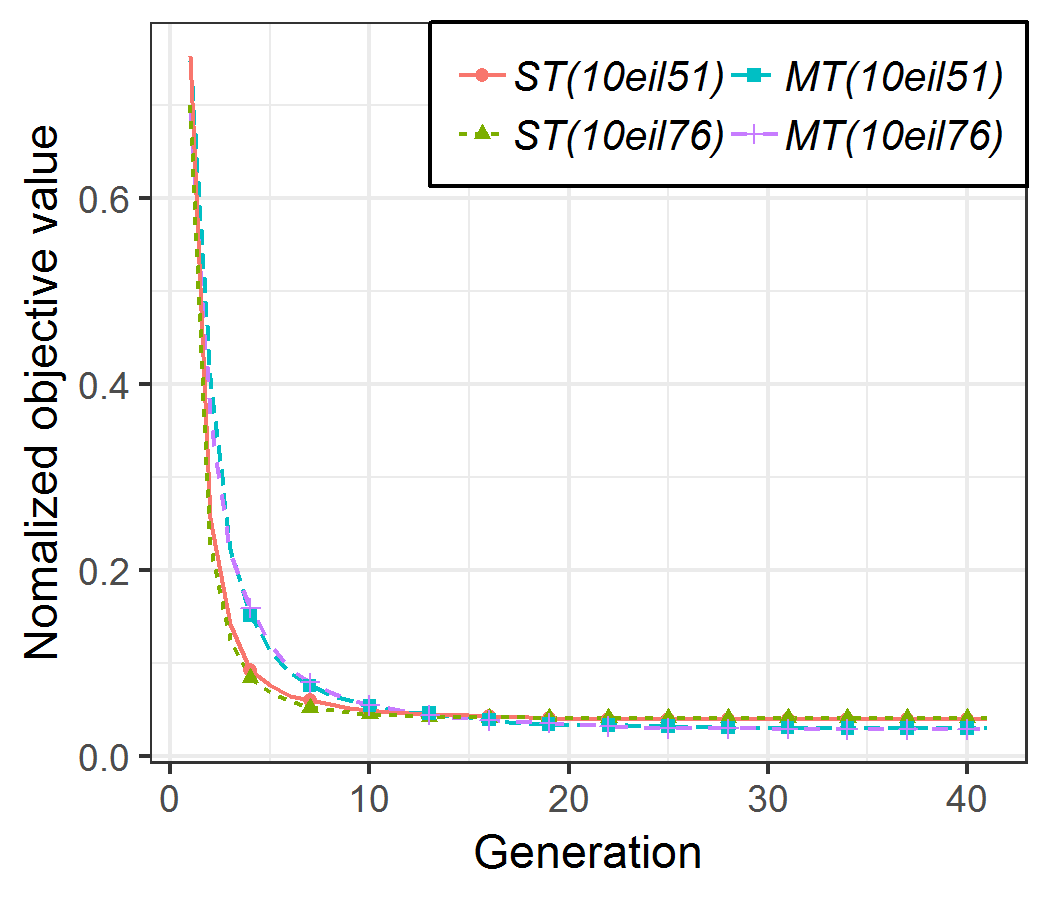}
		\caption{}
		\label{fig:ConvergencyEachTask-a}
	\end{subfigure}
	\begin{subfigure}{1\linewidth}
		\centering
		\includegraphics[scale=\scalefigure]{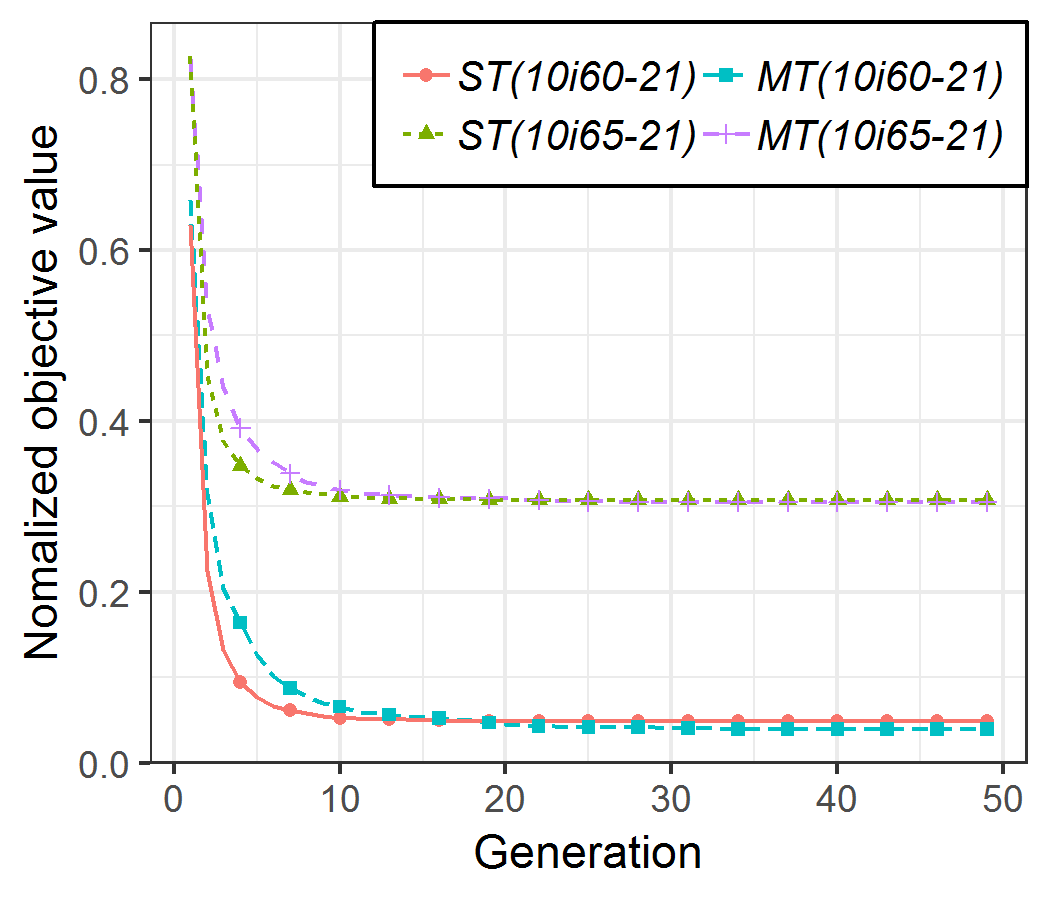}
		\caption{}
		\label{fig:ConvergencyEachTask-b}
	\end{subfigure}
	\begin{subfigure}{1\linewidth}
		\centering
		\includegraphics[scale=\scalefigure]{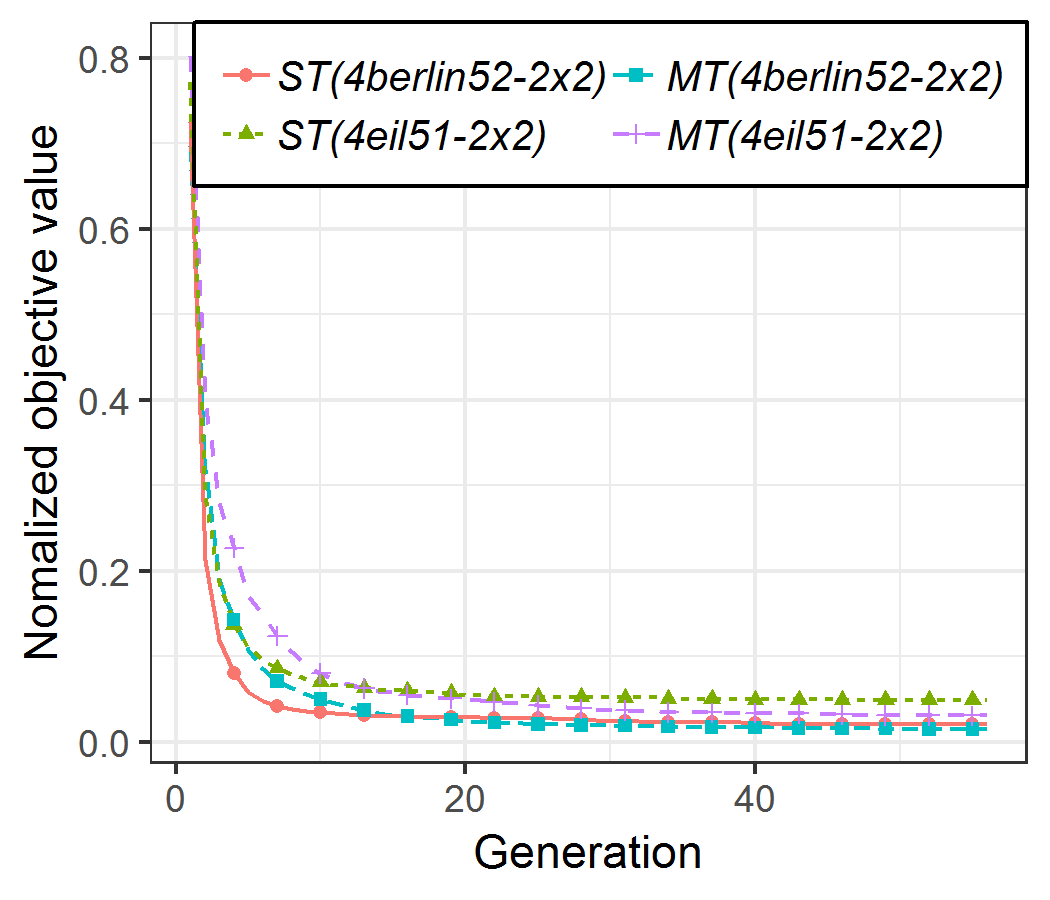}
		\caption{}
		\label{fig:ConvergencyEachTask-c}
	\end{subfigure}
	\caption{Comparing convergence trends of $\tilde{f_1}$ and $\tilde{f_2}$ in multi-tasks and serial single task for instances 10eil51 and 10eil76 in Type 1;  instances 10i60-21 and 10i65-21 in Type 5; instances 4berlin52-2x2 and 4eil51-2x2 in Type 6.}
	\label{fig:ConvergenceTrendForEachTask}
\end{figure}

The major convergence trends of those algorithms in Figure~\ref{fig:ConvergenceTrends} is that MT converges slower than ST for initial generations but MT surpasses ST in later generations which means that the implicit genetic transferring among tasks in evolutionary multitasking paradigm improves the convergence speed of MT in comparison with ST%; secondly MT converges faster than ST in all generations which mean that multi-tasks keep the main features of \gls{ltga}. 

Figure~\ref{fig:ConvergenceTrendForEachTask} provides insight into the improved performance as a consequence of MT. The figure depicts the convergence trends corresponding to each individual task, which is somewhat similar to that of MT and ST in Figure~\ref{fig:ConvergenceTrends} when the convergence rate of each task in MT is better than the corresponding task in ST in later generations.

%% file: Table_Data/DeceptiveTrapSetup-1.tex
%\begin{table*}[htbp]
%  \centering
%  \caption{mk-DTF Instances and Parameters for Evaluating \glsentryshort{mfltga} and \glsentryshort{ltga}}
%    \begin{tabular}{l r r r r r r r r r r r r r r r r r r}
%    \toprule
%    k     & 3     & 3     & 3     & 3     & 3     & 3     & 4     & 4     & 4     & 4     & 4     & 4     & 5     & 5     & 5     & 5     & 5     & 5 \\
%    \cmidrule(l{3pt}r{3pt}){1-1} \cmidrule(l{3pt}r{3pt}){2-7} \cmidrule(l{3pt}r{3pt}){8-13} \cmidrule(l{3pt}r{3pt}){14-19}
%    m     & 5     & 10    & 15    & 20    & 25    & 30    & 5     & 10    & 15    & 20    & 25    & 30    & 5     & 10    & 15    & 20    & 25    & 30 \\
%    \addlinespace
%    Problem length & 15    & 30    & 45    & 60    & 75    & 90    & 20    & 40    & 60    & 80    & 100   & 120   & 25    & 50    & 75    & 100   & 125   & 150 \\
%    \addlinespace
%    Population size & 128   & 128   & 128   & 128   & 128   & 128   & 128   & 128   & 128   & 128   & 128   & 128   & 256   & 256   & 256   & 256   & 256   & 256 \\
%    \bottomrule
%    \end{tabular}%
%  \label{tab:mk-dtf_instances}%
%\end{table*}%

\begin{table*}[htbp]
	\centering
	\caption{mk-DTF Instances and Parameters for Evaluating \glsentryshort{mfltga} and \glsentryshort{ltga}}
	\begin{tabular}{l r r r r r r r r r}
		\toprule
		k     & 3     & 3     & 3     & 3     & 3     & 3     & 4     & 4     & 4   \\
		\cmidrule(l{3pt}r{3pt}){1-1} \cmidrule(l{3pt}r{3pt}){2-7} \cmidrule(l{3pt}r{3pt}){8-10}
		m     & 5     & 10    & 15    & 20    & 25    & 30    & 5     & 10    & 15   \\
		\addlinespace
		Problem length & 15    & 30    & 45    & 60    & 75    & 90    & 20    & 40    & 60   \\
		\addlinespace
		Population size & 128   & 128   & 128   & 128   & 128   & 128   & 128   & 128   & 128  \\
		\hline
		k    & 4     & 4     & 4     & 5     & 5     & 5     & 5     & 5     & 5 \\
		\cmidrule(l{3pt}r{3pt}){1-1} \cmidrule(l{3pt}r{3pt}){2-7} \cmidrule(l{3pt}r{3pt}){8-10}
		m    & 20    & 25    & 30    & 5     & 10    & 15    & 20    & 25    & 30 \\
		\addlinespace
		Problem length & 80    & 100   & 120   & 25    & 50    & 75    & 100   & 125   & 150 \\
		\addlinespace
		Population size  & 128   & 128   & 128   & 256   & 256   & 256   & 256   & 256   & 256 \\
		\bottomrule
	\end{tabular}%
	\label{tab:mk-dtf_instances}%
\end{table*}%

%% file: Table_Data/ResultsDeceptiveTrap-2.tex
% Table generated by Excel2LaTeX from sheet 'TongHop_Paper'
\begin{table}[htbp]
  \centering
  \caption{\glsentryshort{mk-dtf} Results Obtained By \glsentryshort{mfltga} and \glsentryshort{ltga}}.
    \begin{tabular}{ r r r r r r r }
    \toprule
    \multicolumn{2}{c }{} & \multicolumn{2}{c }{\textbf{\glsentryshort{ltga}}} & \multicolumn{2}{c }{\textbf{\glsentryshort{mfltga}}} &  \\
    \cmidrule(l{3pt}r{3pt}){3-4} \cmidrule(l{3pt}r{3pt}){5-6}
    \textbf{k} & \textbf{m} & \textbf{Num.Opt} & \textbf{Num.Evals} & \textbf{Num.Opt} & \textbf{Num.Evals} & \textbf{P.Imp} \\
    \midrule
    3     & 5     & 10    &     4,228.0  & 10    & 2783.2 & 34.2\% \\
    
    3     & 10    & 10    &   13,664.8  & 10    & 8908.8 & 34.8\% \\
    
    3     & 15    & 10    &   24,552.0  & 10    & 13780.8 & 43.9\% \\
    
    3     & 20    & 10    &   34,007.6  & 10    & 23434.8 & 31.1\% \\
    
    3     & 25    & 10    &   52,244.0  & 10    & 28179.2 & 46.1\% \\
    
    3     & 30    & 10    &   57,387.2  & 10    & 39765.2 & 30.7\% \\
    \addlinespace
    4     & 5     & 10    & 7706.4 & 10    & 6642.4 & 13.8\% \\
    
    4     & 10    & 10    & 25615.2 & 10    & 23587.2 & 7.9\% \\
    
    4     & 15    & 10    & 39010.8 & 10    & 38043.2 & 2.5\% \\
    
    4     & 20    & 10    & 64053.2 & 10    & 61209.2 & 4.4\% \\
    
    4     & 25    & 10    & 85021.2 & 10    & 66013.2 & 22.4\% \\
    
    4     & 30    & 10    & 111479.2 & 10    & 95200.0 & 14.6\% \\
    
    \addlinespace
    5     & 5     & 10    & 20342.4 & 10    & 13900.8 & 31.7\% \\
    
    5     & 10    & 10    & 66150.0 & 10    & 49568.4 & 25.1\% \\
    
    5     & 15    & 10    & 123994.4 & 10    & 65327.2 & 47.3\% \\
    
    5     & 20    & 10    & 176378.4 & 9     & 145569.6 & 17.5\% \\
    
    5     & 25    & 10    & 231582.4 & 10    & 128116.8 & 44.7\% \\
    
    5     & 30    & 10    & 284411.2 & 10    & 196322.4 & 31.0\% \\
    \bottomrule
    \multicolumn{7}{p{8cm}}{\begin{scriptsize} Num.Opt: The number of optimal solutions are found. \end{scriptsize} }\\
    \multicolumn{7}{p{8cm}}{\begin{scriptsize} Num.Eval: The Average number of evaluations to success.  \end{scriptsize} }\\
    \multicolumn{7}{p{9cm}}{\begin{scriptsize} P.Imp: The percentage of differences between the average number of evaluations to success. \end{scriptsize} }\\
    \end{tabular}%
  \label{tab:ResultsDeceptiveTrap}%
\end{table}%

%% file: Table_Data/SumeryResults.tex
% Table generated by Excel2LaTeX from sheet 'SoSanhCEC2018v1'
\begin{table*}[htbp]
  \centering
  \caption{Comparison of results obtained by \gls{mfltga} and the existing algorithms.}
    \begin{tabular}{ c p{4cm} r r r r}
	\toprule	
    \multirow{2}[4]{*}{\textbf{Algorithm}} &  & \textbf{Type 1} & \textbf{Type 5} & \textbf{Type 6} & \textbf{Total}\\
	\cmidrule{2-6}          
	&   Number of instances in a Type    & 18    & 14    & 20 & 52 \\
    \midrule
    \multirow{2}[4]{*}{C-MFEA} 
    & Number of instances on which MF-LTGA outperformed C-MFEA & 13    & 14    & 12 & 39\\
	\addlinespace         
	& Maximum PI(MF-LTGA, C-MFEA) & 36.30\% & 25.90\% & 32.10\% &\\
    \midrule
    \multirow{2}[4]{*}{E-MFEA} & Number of instances on which MF-LTGA outperformed E-MFEA & 9     & 14    & 15 & 38 \\
	\addlinespace        
	 & Maximum PI(MF-LTGA, E-MFEA) & 28.20\% & 29.30\% & 34.20\% &\\
    \midrule
    \multirow{2}[4]{*}{LTGA} 
    & Number of instances on which MF-LTGA outperformed LTGA & 13    & 10    & 17 & 42\\
	\addlinespace          
	& Maximum PI(MF-LTGA, LTGA) & 2.50\% & 2.70\% & 3.30\% &\\
    \bottomrule
    \end{tabular}%
  \label{tab:SummaryResults}%
\end{table*}%

%% file: Sections/Sec_Conclusion.tex
This paper introduced a mechanism for combining \gls{ltga} and \gls{mfo}. The novel algorithm kept the main features of both \gls{ltga} and \gls{mfo}, and descripted new methods for building Linkage Tree Model, Assortative Mating and Crossover Operator. The experimental results show that the newly proposed algorithms were more effective in solving the canonical \gls{clustp} and \gls{dtf} compared with some other existing meta-heuristics.

Several theoretical aspects of the \gls{mfltga} will be investigated in more detail. In the future, we will focus on methods for constructing only one Linkage Tree Model for all tasks.

%% file: Table_Data/ResultsType1.tex
% Table generated by Excel2LaTeX from sheet 'SoSanhCEC2018v2'
\begin{table*}[htbp]
  \centering
  \caption{Results Obtained By E-MFEA, C-MFEA, LTGA and MF-LTGA on Instances In Type 1}
    \begin{tabular}{ l r r r r r r r r}
    \toprule
    \multicolumn{1}{c}{\multirow{2}[4]{*}{\textbf{Instances}}} & \multicolumn{2}{c}{\textbf{E-MFEA}} & \multicolumn{2}{c}{\textbf{C-MFEA}} & \multicolumn{2}{c}{\textbf{LTGA}} & \multicolumn{2}{c}{\textbf{MF-LTGA}} \\
    
    \cmidrule(l{3pt}r{3pt}){2-3} \cmidrule(l{3pt}r{3pt}){4-5} \cmidrule(l{3pt}r{3pt}){6-7} \cmidrule(l{3pt}r{3pt}){8-9}

	& \multicolumn{1}{c}{\textbf{BF}} & \multicolumn{1}{c}{\textbf{Avg}} & \multicolumn{1}{c}{\textbf{BF}} & \multicolumn{1}{c}{\textbf{Avg}} & \multicolumn{1}{c}{\textbf{BF}} & \multicolumn{1}{c}{\textbf{Avg}} & \multicolumn{1}{c}{\textbf{BF}} & \multicolumn{1}{c}{\textbf{Avg}} \\
	
	\midrule
    10berlin52 & 46185.8 & 46707.8 & 48569.8 & 48569.8 & 44269.0 & 45071.3 & 44331.3 & \textcolor[rgb]{ 1,  0,  0}{44869.7} \\
     
    10eil51 & 2008.3 & 2039.4 & 1891.7 & 1891.7 & 1787.5 & 1861.2 & 1726.2 & \textcolor[rgb]{ 1,  0,  0}{1829.4} \\
     
    10eil76 & 2775.4 & 2973.3 & 2489.5 & \textit{2489.5} & 2296.0 & 2505.6 & 2324.7 & \textcolor[rgb]{ 1,  0,  0}{2442.3} \\
     
    10kroB100 & 198181.6 & 218275.1 & 170695.2 & 170695.2 & 151186.8 & 158277.4 & 147780.3 & \textcolor[rgb]{ 1,  0,  0}{156804.6} \\
    \addlinespace
    10pr76 & 643903.6 & 665835.1 & 632704.5 & 632704.5 & 551924.2 & 579248.6 & 533900.2 & \textcolor[rgb]{ 1,  0,  0}{566138.2} \\
     
    10rat99 & 10427.6 & 10792.8 & 8937.1 & 8937.1 & 7778.2 & 8412.0 & 7904.5 & \textcolor[rgb]{ 1,  0,  0}{8323.5} \\
     
    15eil51 & 1662.9 & \textit{\textbf{1781.4}} & 1922.0 & \textit{\textbf{1922.0}} & 1984.6 & 2130.4 & 2019.9 & \textcolor[rgb]{ 1,  0,  0}{2099.7} \\
     
    15eil76 & 3349.0 & \textit{\textbf{3402.8}} & 3773.0 & 3773.0 & 3305.0 & 3496.0 & 3318.9 & \textcolor[rgb]{ 1,  0,  0}{3436.6} \\
     
    15pr76 & 772173.1 & \textit{\textbf{787889.0}} & 833734.0 & \textit{833734.0} & 814795.5 & 835549.4 & 808406.1 & \textcolor[rgb]{ 1,  0,  0}{829608.7} \\
    \addlinespace
    15st70 & 4972.1 & 5117.7 & 5171.8 & 5171.8 & 4362.2 & 4522.5 & 4335.8 & \textcolor[rgb]{ 1,  0,  0}{4435.4} \\
     
    25eil101 & 5192.4 & 5248.4 & 6852.3 & 6852.3 & 4828.9 & 4983.8 & 4852.9 & \textcolor[rgb]{ 1,  0,  0}{4950.5} \\
     
    25kroA100 & 164038.4 & \textit{\textbf{167528.9}} & 266798.7 & 266798.7 & 163607.8 & 171118.9 & 165696.9 & \textcolor[rgb]{ 1,  0,  0}{170175.5} \\
     
    25lin105 & 106500.2 & \textit{\textbf{107524.2}} & 182650.4 & 182650.4 & 133620.8 & 141926.3 & 136782.0 & \textcolor[rgb]{ 1,  0,  0}{139844.5} \\
     
    25rat99 & 9234.7 & 9375.8 & 12931.7 & 12931.7 & 8072.6 & 8417.5 & 8213.1 & \textcolor[rgb]{ 1,  0,  0}{8446.7} \\
    \addlinespace
    50eil101 & 3978.2 & \textit{\textbf{3991.2}} & 9461.1 & \textbf{9461.1} & 7985.5 & 8150.7 & 12356.2 & 12378.8 \\
     
    50kroA100 & 173626.5 & \textit{\textbf{176321.3}} & 451952.5 & \textbf{451952.5} & 289992.4 & 298047.5 & 525968.7 & 527328.5 \\
     
    50kroB100 & 138019.5 & \textit{\textbf{138849.1}} & 450713.7 & \textbf{450713.7} & 236715.1 & 245304.7 & 538061.6 & 538991.1 \\
     
    50lin105 & 147967.2 & \textit{\textbf{148332.6}} & 309399.9 & \textbf{309399.9} & 226340.0 & 232625.7 & 392391.3 & 392750.7 \\
    \bottomrule
    \end{tabular}%
  \label{tab:ResultsType1}%
\end{table*}%

%% file: Table_Data/ResultsType5.tex
% Table generated by Excel2LaTeX from sheet 'SoSanhCEC2018v2'
\begin{table*}[htbp]
  \centering
  \caption{Results Obtained By E-MFEA, C-MFEA, LTGA and MF-LTGA on Instances In Type 5}
    \begin{tabular}{ l r r r r r r r r}
     \toprule
    \multicolumn{1}{c}{\multirow{2}[4]{*}{\textbf{Instances}}} & \multicolumn{2}{c}{\textbf{E-MFEA}} & \multicolumn{2}{c}{\textbf{C-MFEA}} & \multicolumn{2}{c}{\textbf{LTGA}} & \multicolumn{2}{c}{\textbf{MF-LTGA}} \\
    
	\cmidrule(l{3pt}r{3pt}){2-3} \cmidrule(l{3pt}r{3pt}){4-5} \cmidrule(l{3pt}r{3pt}){6-7} \cmidrule(l{3pt}r{3pt}){8-9}   
	      
	& \multicolumn{1}{c}{\textbf{BF}} & \multicolumn{1}{c}{\textbf{Avg}} & \multicolumn{1}{c}{\textbf{BF}} & \multicolumn{1}{c}{\textbf{Avg}} & \multicolumn{1}{c}{\textbf{BF}} & \multicolumn{1}{c}{\textbf{Avg}} & \multicolumn{1}{c}{\textbf{BF}} & \multicolumn{1}{c}{\textbf{Avg}} \\
    \midrule
     
    10i120-46 & 119168.5 & 120920.6 & 105754.4 & 125137.1 & 98742.1 & 101766.7 & 97389.6 & 101961.4 \\
      
    10i45-18 & 26065.2 & 27515.4 & 26942.8 & 32663.5 & 24209.2 & 25404.1 & 23739.6 & \textcolor[rgb]{ 1,  0,  0}{24727.7} \\
      
    10i60-21 & 43125.9 & 45389.1 & 37640.0 & 45427.1 & 34424.5 & 37629.5 & 34424.5 & \textcolor[rgb]{ 1,  0,  0}{37346.9} \\
      
    10i65-21 & 46456.5 & 48420.7 & 41053.9 & 49824.3 & 39137.4 & 41928.4 & 39511.7 & 41965.8 \\
    \addlinespace
    10i70-21 & 49875.8 & 51532.2 & 41892.8 & 55760.2 & 38714.6 & 42618.6 & 39506.1 & \textcolor[rgb]{ 1,  0,  0}{41476.8} \\
      
    10i90-33 & 61567.6 & 64955.9 & 55361.9 & 65493.1 & 54402.0 & 57278.2 & 55186.4 & 57545.4 \\
      
    5i45-18 & 20042.8 & 22345.5 & 15511.5 & 17007.0 & 14986.9 & 16063.1 & 15068.7 & \textcolor[rgb]{ 1,  0,  0}{15792.3} \\
      
    5i60-21 & 35099.9 & 36474.5 & 29797.9 & 34613.0 & 29263.8 & 31129.9 & 29113.7 & \textcolor[rgb]{ 1,  0,  0}{30500.8} \\
      
    5i65-21 & 33648.4 & 35663.9 & 31517.4 & 34235.2 & 31238.8 & 33537.1 & 31829.7 & \textcolor[rgb]{ 1,  0,  0}{33180.1} \\
    \addlinespace
    5i70-21 & 43816.9 & 49519.0 & 35746.0 & 39302.4 & 35096.6 & 38044.7 & 35706.8 & \textcolor[rgb]{ 1,  0,  0}{37763.8} \\
      
    5i75-22 & 37992.7 & 40668.9 & 34867.3 & 38705.1 & 35700.9 & 38396.4 & 35361.8 & \textcolor[rgb]{ 1,  0,  0}{37813.2} \\
      
    5i90-33 & 62701.2 & 65622.1 & 53230.6 & 55888.2 & 53195.0 & 56485.3 & 53592.2 & 56604.4 \\
      
    7i60-21 & 44669.6 & 46337.4 & 37690.6 & 41532.3 & 36692.9 & 40318.7 & 37186.7 & \textcolor[rgb]{ 1,  0,  0}{40062.7} \\
      
    7i65-21 & 45237.3 & 47211.2 & 35878.8 & 40222.5 & 36230.3 & 38858.5 & 35924.4 & \textcolor[rgb]{ 1,  0,  0}{38116.1} \\
    \bottomrule
    \end{tabular}%
  \label{tab:ResultsType5}%
\end{table*}%

%% file: Table_Data/ResultsType6.tex
% Table generated by Excel2LaTeX from sheet 'SoSanhCEC2018v2'
\begin{table*}[htbp]
  \centering
  \caption{Results Obtained By E-MFEA, C-MFEA, LTGA and MF-LTGA on Instances In Type 6}
    \begin{tabular}{ l r r r r r r r r}
    \toprule
    \multicolumn{1}{c}{\multirow{2}[4]{*}{\textbf{Instances}}} & \multicolumn{2}{c}{\textbf{E-MFEA}} & \multicolumn{2}{c}{\textbf{C-MFEA}} & \multicolumn{2}{c}{\textbf{LTGA}} & \multicolumn{2}{c}{\textbf{MF-LTGA}} \\
	
	 \cmidrule(l{3pt}r{3pt}){2-3} \cmidrule(l{3pt}r{3pt}){4-5} \cmidrule(l{3pt}r{3pt}){6-7} \cmidrule(l{3pt}r{3pt}){8-9}
	 
	& \multicolumn{1}{c}{\textbf{BF}} & \multicolumn{1}{c}{\textbf{Avg}} & \multicolumn{1}{c}{\textbf{BF}} & \multicolumn{1}{c}{\textbf{Avg}} & \multicolumn{1}{c}{\textbf{BF}} & \multicolumn{1}{c}{\textbf{Avg}} & \multicolumn{1}{c}{\textbf{BF}} & \multicolumn{1}{c}{\textbf{Avg}} \\
    \midrule
    10berlin52-2x5 & 31659.1 & \em{\textbf{33590.6}} & -     & -     & 35643.0 & 37342.1 & 36449.8 & \textcolor[rgb]{ 1,  0,  0}{37222.8} \\
       
    12eil51-3x4 & 1922.1 & 1960.9 & 2185.0 & 2691.5 & 1782.7 & 1867.3 & 1749.8 & \textcolor[rgb]{ 1,  0,  0}{1826.5} \\
       
    12eil76-3x4 & 3352.2 & 3449.3 & 3065.7 & 3896.1 & 2852.6 & 2938.8 & 2752.4 & \textcolor[rgb]{ 1,  0,  0}{2892.6} \\
       
    12pr76-3x4 & 664562.6 & \em{\textbf{685351.9}} & -     & -     & 720063.9 & 767947.9 & 713144.0 & \textcolor[rgb]{ 1,  0,  0}{751440.1} \\
       
    12st70-3x4 & 4750.9 & 4795.7 & -     & -     & 4265.6 & 4391.7 & 4619.6 & 4753.3 \\
    \addlinespace
    15pr76-3x5 & 601015.9 & \em{\textbf{623645.7}} & -     & -     & 763415.0 & 838763.6 & 778422.9 & \textcolor[rgb]{ 1,  0,  0}{815522.3} \\
       
    16eil51-4x4 & 1371.3 & \em{\textbf{1425.9}} & -     & -     & 1779.3 & 1843.7 & 1775.8 & \textcolor[rgb]{ 1,  0,  0}{1814.1} \\
       
    16eil76-4x4 & 2314.0 & \em{\textbf{2374.9}} & -     & -     & 2705.3 & 2836.7 & 2569.5 & \textcolor[rgb]{ 1,  0,  0}{2772.3} \\
       
    16lin105-4x4 & 179729.7 & 179729.7 & -     & -     & 160926.8 & 170624.9 & 161710.3 & \textcolor[rgb]{ 1,  0,  0}{166233.5} \\
       
    16st70-4x4 & 3560.4 & 3560.4 & -     & -     & 3248.9 & 3481.6 & 3235.8 & \textcolor[rgb]{ 1,  0,  0}{3369.7} \\
    \addlinespace
    4berlin52-2x2 & 35413.1 & 37121.5 & 23635.3 & 24751.0 & 23287.9 & 24664.1 & 23509.9 & \textcolor[rgb]{ 1,  0,  0}{24431.5} \\
       
    4eil51-2x2 & 2545.3 & 2641.1 & 1909.5 & 2053.9 & 1934.7 & 2058.0 & 1911.6 & \textcolor[rgb]{ 1,  0,  0}{2010.3} \\
       
    4eil76-2x2 & 4319.3 & 4517.2 & 2949.1 & 3179.9 & 2977.2 & 3231.6 & 3023.5 & \textcolor[rgb]{ 1,  0,  0}{3198.8} \\
       
    4pr76-2x2 & 688228.2 & 762880.2 & 446862.4 & 480043.8 & 450274.6 & 513032.1 & 461545 & 546392.3 \\
       
    6pr76-2x3 & 741847.3 & 771563.7 & 656978.3 & 736743.5 & 661240.6 & 703446.5 & 655261.7 & \textcolor[rgb]{ 1,  0,  0}{694048.9} \\
    \addlinespace
    6st70-2x3 & 3880.5 & 4057.0 & 3508.1 & 4244.1 & 3501.1 & 3764.2 & 3522.32548 & \textcolor[rgb]{ 1,  0,  0}{3729.8} \\
       
    9eil101-3x3 & 4281.2 & 4585.0 & 3320.3 & 4345.8 & 3292.9 & 3545.5 & 3334.0 & \textcolor[rgb]{ 1,  0,  0}{3501.9} \\
       
    9eil51-3x3 & 2127.0 & 2182.7 & 2106.5 & 2630.8 & 1954.5 & 2073.5 & 1942.8 & \textcolor[rgb]{ 1,  0,  0}{2036.8} \\
       
    9eil76-3x3 & 3599.5 & 3730.8 & 3401.8 & 4048.0 & 3059.1 & 3253.1 & 3091.1 & \textcolor[rgb]{ 1,  0,  0}{3225.8} \\
       
    9pr76-3x3 & 713966.7 & 749030.6 & 642796.6 & 783056.1 & 576233.6 & 599439.8 & 576233.6 & 599439.8 \\
    \bottomrule
    \end{tabular}%
  \label{tab:ResultsType6}%
\end{table*}%